\documentclass[conference]{IEEEtran}
\IEEEoverridecommandlockouts
\usepackage{cite}
\usepackage{amsmath,amssymb,amsfonts}
\usepackage{algorithmic}
\usepackage{graphicx}
\usepackage{textcomp}
\usepackage{float}
\usepackage{soul}
\usepackage{multirow,color}
\usepackage{array} 
\usepackage{tikz,balance}
\usepackage{url}
\usepackage{subcaption}
\usepackage{enumitem,xcolor,booktabs,colortbl}
\def\BibTeX{{\rm B\kern-.05em{\sc i\kern-.025em b}\kern-.08em
    T\kern-.1667em\lower.7ex\hbox{E}\kern-.125emX}}

\begin{document}
\title{Generating Game Levels of Diverse Behaviour Engagement
}

\author{
  \IEEEauthorblockN{Keyuan Zhang\IEEEauthorrefmark{1}\IEEEauthorrefmark{2}, Jiayu Bai\IEEEauthorrefmark{1}\IEEEauthorrefmark{2}, Jialin Liu\IEEEauthorrefmark{2}\IEEEauthorrefmark{1}}
  \IEEEauthorblockA{
    \IEEEauthorrefmark{1} \textit{Research Institute of Trustworthy Autonomous System}\\
    \textit{Southern University of Science and Technology (SUSTech)}}
    \IEEEauthorblockA{\IEEEauthorrefmark{2}\textit{Guangdong Provincial Key Laboratory of Brain-inspired Intelligent Computation}\\
    \textit{Department of Computer Science and Engineering} \\
    \textit{Southern University of Science and Technology (SUSTech)}\\
    Shenzhen, China
  }
}

\maketitle

\begin{abstract}
Recent years, there has been growing interests in experience-driven procedural level generation. Various metrics have been formulated to model player experience and help generate personalised levels. In this work, we question whether experience metrics can adapt to agents with different personas. We start by reviewing existing metrics for evaluating game levels. Then, focusing on platformer games, we design a framework integrating various agents and evaluation metrics. Experimental studies on \emph{Super Mario Bros.} indicate that using the same evaluation metrics but agents with different personas can generate levels for particular persona. It implies that, for simple games, using a game-playing agent of specific player archetype as a level tester is probably all we need to generate levels of diverse behaviour engagement.

\end{abstract}


\section{Introduction}\label{sec:introduction}

Procedural content generation (PCG), aiming at generating contents algorithmatically, has shown its effectiveness in generating various types of game contents~\cite{togelius2013procedural,Summerville2018pcgml,risi2020increasing,liu2021dlpcg}. As the rapid development and applications of video games in different fields (e.g., serious games for autistic treatment) and expansion of player demography, game content tends to satisfy prospective players with different preference and experience~\cite{Yannakakis2011Experience}. More and more studies start to focus on experience-driven procedural content generation (EDPCG). To generate personalised contents, it is necessary to define evaluation metrics that can predict player experience on the evaluated contents. Basically, an evaluation metric can be seen as a function whose input is the gameplay data or content features that are controllable by game designers and its output is the quality of the content. This mapping is often mathematically formulated or learnt by machine learning models~\cite{Pedersen2010Modeling,gravina2019procedural}. For instance, Pedersen \emph{et al.}~\cite{Pedersen2010Modeling} trained a model to predict several key affective states with a combination of gameplay features and controllable content features. 

EDPCG can be applied to generate contents for general players or a particular player persona. EDPCG for general players exists in many forms of games, such as platformer games~\cite{shu2021edrl}, dungeon games~\cite{alvarez2018assessing}, puzzle games~\cite{Sarkar2020SequentialSL} and role playing games~\cite{Nam2019GenerationOD}. Besides, Some work categorised players into different groups according to the behaviour preference (personas) and try to generate content for specific groups\cite{Stammer2015player, yu2011personalized}. The common procedure of generating content for specific persona has two steps, identifying which profile the player belongs to, and then changing or adapting the generators correspondingly.

What if the same content evaluation metrics are used for different personas? There exists at least two advantages. (i) It is not necessary to design different content evaluation metrics for different personas. (ii) If the playing style changes, the generated content can also change since all the different playing styles share the same evaluation metrics. To our best knowledge, no work has ever considered general evaluation metrics for different player personas. Fernandes \emph{et al.}~\cite{fernandes2021adapting} proposed a level generation approach to adapt to four rule-based persona agents over three different experience metrics respectively. However, the work~\cite{fernandes2021adapting} did not show the difference between the levels generated for different personas. This is important because it reflects whether the evaluation metrics can capture the preference of personas and affect the generation of levels.

Motivated by the above, two questions are raised in this work.
How different the generated levels are if using the same evaluation metric but different evaluation agents? Furthermore, considering a level generated with a given evaluation agent, will a player that has similar persona to the evaluation agent gain more engagement compared with players that have another persona? In this paper, we attempt to answer the above questions with level generation and conduct case studies on \emph{Super Mario Bros.} (SMB), a benchmark platformer game for procedural level generation.

The main contributions of this work are as follows\footnote{Code of this paper is available on Github: \url{https://github.com/SUSTechGameAI/EngagementMetrics}}.
First, existing metrics for evaluating game levels (not particularly for platformer game levels) are reviewed. Then, we propose a framework that is capable of generating levels of diverse behaviour engagement. The framework is implemented with three agents of different personas and four evaluation metrics considering the number, difficulty and diversity of events triggered by agents during game-playing.
The aforementioned metrics and agents are integrated to the popular MarioGan~\cite{Volz2018Evolving} framework as fitness functions of an evolutionary algorithm which searches in the latent space of the level generator of MarioGan.
Levels obtained with different combinations of evaluation metrics and agents are evaluated by their contents and gameplay data of different agents in order to verify their difference and whether they can adapt to agents used during level generation.

The rest of this paper is organised as follows. Section \ref{sec:related} reviews existing metrics for evaluating game levels. The motivation and design of our framework, details of an implementation of our framework for generating SMB levels, our evaluation metrics and agents are provided in Section \ref{sec:study}. Section \ref{sec:xp} presents the corresponding experimental studies and analysis.
Section \ref{sec:conclusion} concludes and discusses some future directions.


\section{Related Work}\label{sec:related}

\begin{table*}[ht]
    \centering
    \caption{Factors considered when evaluating levels and research on level generation for different players'/agents' personas.}
    \label{tab:research}
    \begin{tabular}{p{0.6cm}|c|c|c|p{3.5cm}|p{3.5cm}|p{3.5cm}}
    \toprule
    \multirow{2}{0.6cm}{\textbf{Work}} & \multicolumn{3}{c}{\textbf{Evaluation factors}} & \multirow{2}{3.5cm}{\textbf{Whether categorize players/agents (Criterion)}} & \multirow{2}{3.5cm}{\textbf{Apply same evaluation metrics for different groups}} & \multirow{2}{3.5cm}{\textbf{Compare generated levels for different groups}} \\
     & content & gameplay & feedback &  &  &   \\
    \hline
    \cite{Yannakakis2009RealTimeGA} & $\checkmark$ & $\checkmark$ & $\checkmark$ & - & - & -\\
    \hline
    \cite{Melhrt2021TowardsGM} &  & $\checkmark$ & $\checkmark$ & - & - & -\\
    \hline
    \cite{Pedersen2010Modeling} & $\checkmark$ & $\checkmark$ & $\checkmark$ & - & - & - \\
    \hline
    \cite{Multi-Dimensional} & $\checkmark$ & $\checkmark$ & $\checkmark$ & - & - & - \\ 
    \hline
    \cite{Nam2019GenerationOD} & $\checkmark$ & $\checkmark$ & - & -& - & - \\
    \hline
    \cite{Hald2020ProceduralCG} & $\checkmark$& - & - & - & - & - \\
    \hline
    \cite{Sarkar2020SequentialSL} & $\checkmark$ & $\checkmark$ & - & -& - & - \\
    \hline
    \cite{shu2021edrl} & $\checkmark$ & $\checkmark$ & - & - & - & - \\
    \hline
    \cite{yu2011personalized} & $\checkmark$ & $\checkmark$ & $\checkmark$ & $\checkmark$ (Persona) & - & - \\
    \hline
    \cite{Stammer2015player} & $\checkmark$ & $\checkmark$ & - & $\checkmark$ (Persona) & - & - \\
    \hline
    \cite{Green2018Generating} &  $\checkmark$ & $\checkmark$ & - & $\checkmark$ (Ability) & $\checkmark$ & $\checkmark$ \\
    \hline
    \cite{fernandes2021adapting} & $\checkmark$ & - & - & $\checkmark$ (Persona) & $\checkmark$ & - \\
    \hline
    Self & - & $\checkmark$ & - & $\checkmark$ (Persona) & $\checkmark$ & $\checkmark$ \\
    \bottomrule
    \end{tabular}
\end{table*}

Section \ref{sec:persona} briefly reviews the related work in generating contents for different player (or agent) types.
Focusing on game levels, Section \ref{sec:metrics} summarises and discusses what have been modelled in existing level evaluation metrics.

\subsection{Generating Levels for Different Player / Agent Types}\label{sec:persona}
Personalised level design has been comprehensively summarised in the survey of~\cite{snodgrass2019like}. Here, research work that classifies players into several groups and generate levels are discussed. 
The classification approaches can be based on intuition or machine learning. The work of \cite{Stammer2015player} categorised playing styles as Explorer, Enemy killer, Speed runner and the work of \cite{fernandes2021adapting} defined four rule-based agent.
Yu and Trawick \cite{yu2011personalized} applied some clustering methods to categorise players and the naive Bayesian approach to identify playing styles. 
As shown in Table \ref{tab:research}, \cite{Stammer2015player} and \cite{yu2011personalized} applied different level evaluation metrics to different playing styles, while some work~\cite{Green2018Generating,fernandes2021adapting} used the same evaluation metrics but different agents for simulating the game.


\subsection{Evaluating Levels}\label{sec:metrics}
Table \ref{tab:research} also summarises the factors considered when designing a level evaluation metric in each related work, including the level content itself, game play data collected during the games played by players or agents and players' feedback about their feelings.

The level evaluation metrics can be modelled by expert knowledge or machine learning methods. Most work considering players' feedback trained a model that can predict ``fun'' degree of the levels based on the level content and (or) gameplay data~\cite{Yannakakis2009RealTimeGA, Melhrt2021TowardsGM, Pedersen2010Modeling, yu2011personalized}. The work of \cite{Multi-Dimensional} used the feedback of players during the game to adaptively change the challenge of the game. Although the work focusing on player modelling are not listed in Table \ref{tab:research}, they can also help level evaluation by imitating the decision making or players' playing style~\cite{Zhou2020Discovering, Holmgrd2016EvolvingMO, Drachen2009PlayerMU, Melhrt2019YourGS}.

\section{Approach and Case study with SMB}\label{sec:study}




In this work, we mainly consider the gameplay data collected during the games because we want to focus on whether the evaluation metrics can be sensitive to agents with different behaviour preference. We also expect to see if the generation of levels will change for different agents when only the gameplay data is used to evaluate levels.

\begin{figure}[hbp]
\centering
\includegraphics[width=0.45\textwidth]{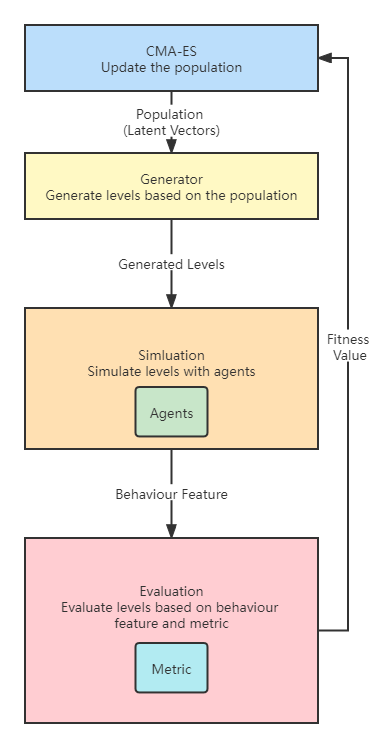}
\caption{\label{framework}Our framework built on MarioGan~\cite{Volz2018Evolving}. 
}
\end{figure}

Our proposed framework is illustrated in Fig. \ref{framework}. It is a general framework that can be extended to many games. Basically, it is a search-based approach with a generator to create new candidate levels, a simulator to generate gameplay data on the candidate levels and some evaluation metrics to evaluate the quality of the candidate levels based on the gameplay data.

Our framework is based on MarioGan~\cite{Volz2018Evolving}, but differs from it as follows. In MarioGan~\cite{Volz2018Evolving}, only one agent was used to simulate the generated levels and only the level completion rate and number of jumps by the agent are considered when evaluating levels. We extend it by considering various agents of different playing styles as level evaluation agents. Besides the original evaluation metrics, new metrics are designed considering more gameplay data, including the event number, event distribution and ability of agents.

\subsection{Agents of Different Personas}\label{sec:agents}

Inspired by the Bartle's Taxonomy \cite{bartle2004designing}, which categorised players to \textit{Killer}, \textit{Achiever}, \textit{Explorer} and \textit{Socializer}, we design three different SMB agents: \textit{Runner}, \textit{Killer} and \textit{Collector}.
Despite of their common interest in passing the levels, their preference and reaction to different game mechanisms (such as coins and monsters) are different.

Runner, Killer and Collector agents are designed as variations of the winner $A^*$ agent in the 2009 Mario AI Competition~\cite{togelius20102009}.
Its ability of passing levels has been proved to be superior than human players. Different personas are implemented through the design of different heuristic function used for decision-making by each agent.
The personas and corresponding agents are described as follows.

\subsubsection{Runner}
\textit{Runner} represents the players that try to complete a level as fast as possible. Speed run is a common way for players to break the game record and avoid monsters in many games.
Its heuristic function punishes the actions that lead to states which cost longer time to finish, formulated as
\begin{equation}
  cost_r = RemainingTime + TimeElapsed * 0.9,
\end{equation}
where $RemainingTime$ is the estimated time consumed by Mario to finish the level and $TimeElapsed$ is simply the elapsed time.

\subsubsection{Killer}
\textit{Killer} represents the players that prefer to kill all the monsters in a level. It represents an extreme case for players that 
enjoy stomping monsters. Its heuristic function rewards the actions that kill monsters, formulated as
\begin{equation}
cost_k = - KillRate -  GameState,
\end{equation}
where $GameState=1$ if Mario reaches the destination; $GameState=-1$ if Mario falls or is killed by a monster; otherwise, $GameState=0$. 
Introducing $GameState$ is to ensure the priority of completing levels since it is the common goal of all the players. $KillRate$ is the ratio of monster killed by Mario and the total monsters.

\subsubsection{Collector}
\textit{Collector} represents the players that attempt to collect all the coins in a level. It represents achievers~\cite{bartle2004designing} who have fun when collecting all the props and earning additional reward in games. Its heuristic function rewards the actions that collect more coins:
\begin{equation}
cost_c = -  CollectRate -  GameState, 
\end{equation}
where $CollectRate$ is the ratio of collected coins to the total number of coins.


\subsection{Design of Evaluation Metrics}\label{sec:designmetrics}

Five different evaluation metrics are designed to evaluate the quality of levels by the agents' gameplay features. Note that all the metrics are to be minimised in our work. 

\subsubsection{Jump}

The number of jumps is a common metrics used in SMB~\cite{Volz2018Evolving, Pedersen2010Modeling} since \texttt{jump} is a basic operation for players. To be specific, we want to maximise the number of jumps of agents and formulate the following evaluation metric: 
\begin{equation}\label{eqa:jumps}
Jump=
\left\{
\begin{aligned}
&-p  & p < 1, \\
&-p - \#Jumps & p = 1,
\end{aligned}
\right.
\end{equation}
\noindent where $p$ is the fraction of levels that the agent completes. $p$ is to ensure that the level is playable. It is negative because we want to minimise the value.

\subsubsection{Event}
Since jumping can be seen as an event during playing, we use the number of events as the metrics and want to study their difference for generating levels. Table \ref{event} lists all the considered events. The corresponding metric is:
\begin{equation}
Event=
\left\{
\begin{aligned}
&-p  & p < 1, \\
&-p - \#E & p = 1,
\end{aligned}
\right.   
\end{equation}
\noindent where $\#E$ is the number of events occurred.

\begin{table}[htbp]
  \centering
  \caption{Events used for evaluating levels.}
    \begin{tabular}{c|c}
    \toprule
    \textbf{Event type} & \textbf{Description} \\
    \midrule
    Stomp & Stomp on and kill a monster \\
    Fall & Monster falls \\
    Jump & Jump up \\
    Land & Land on the solid block \\
    Collect & Collect a coin \\
    Lose & Fall or be killed by a monster \\
    Win & Complete the level \\
    \bottomrule
    \end{tabular}
  \label{event}
\end{table}

\subsubsection{Fail rate}

Fail rate is used to describe the danger during playing. As an $A^*$ agent, a search-based agent, is used, the ratio of the nodes searched by $A^*$ that cause a lose (denoted as $S_{lose}$) to the total nodes searched by $A^*$ (denoted as $S_{total}$) can be calculated as $fr = \text{\#}S_{lose} / \text{ \#} S_{total}$.
We want to maximise the fail rate of the agent in order to increase the difficulty degree of the levels, formulated as follows:
\begin{equation}
FailRate=
\left\{
\begin{aligned}
&-p  & p < 1, \\
&-p - fr & p = 1.
\end{aligned}
\right.    
\end{equation}

\begin{table*}[htbp]
  \centering
  \caption{The ratio of completion, number of monsters killed, number of coins collected and complete time of three agents tested on the 15 original SMB levels. AVG in ``Monsters killed'' (or ``Coins Collected'') refers to the average ratio of the monsters killed (or coins collected) by agents. $-1$ in ``Time'' means that the agent fails to finish the level. The maximum ratio of completion, monsters killed, coins collected and the minimum completion time are in bold. Some cells are left empty because it is meaningless to calculate the averaged value as the levels are different.}
    \begin{tabular}{c|c|ccccccccccccccc|c}
    \toprule
    \textbf{Level} &   \textbf{Agent}    & \textbf{1}     & \textbf{2}     & \textbf{3}     & \textbf{4}     & \textbf{5}     & \textbf{6}     & \textbf{7}     & \textbf{8}     & \textbf{9}     & \textbf{10}    & \textbf{11}    & \textbf{12}    & \textbf{13}    & \textbf{14}    & \textbf{15}    & \textbf{AVG} \\
    \midrule
    \multirow{3}[2]{*}{Completion} & \textit{Runner} & \textbf{1} & \textbf{1} & 0.53  & \textbf{1} & \textbf{1} & \textbf{0.41} & \textbf{1} & \textbf{1} & \textbf{1} & 0.53  & \textbf{1} & 0.37  & \textbf{1} & \textbf{1} & 0.11  & \multicolumn{1}{c}{0.79} \\
          & \textit{Killer} & \textbf{1} & \textbf{1} & \textbf{0.71} & 0.93  & \textbf{1} & \textbf{0.41} & \textbf{1} & 0.87  & \textbf{1} & \textbf{0.57} & \textbf{1} & \textbf{1} & 0.35  & \textbf{1} & \textbf{0.93} & \multicolumn{1}{c}{\textbf{0.85}} \\
          & \textit{Collector} & \textbf{1} & \textbf{1} & 0.29  & 0.94  & \textbf{1} & 0.4   & \textbf{1} & \textbf{1} & \textbf{1} & 0.52  & \textbf{1} & \textbf{1} & 0.35  & \textbf{1} & \textbf{0.92} & \multicolumn{1}{c}{0.82} \\
    \midrule
    \multirow{4}[2]{*}{Monsters killed} & \textit{Runner} & 1     & 5     & 0     & 8     & 8     & 0     & 0     & 0     & 5     & 0     & 0     & 1     & 0     & 4     & 3.2   & \multicolumn{1}{c}{0.12} \\
          & \textit{Killer} & \textbf{12.8} & \textbf{17} & \textbf{4.6} & \textbf{21.8} & \textbf{20.8} & \textbf{2} & 0     & \textbf{11.6} & \textbf{22} & \textbf{5} & 0     & \textbf{6.2} & 0     & \textbf{7} & \textbf{36} & \multicolumn{1}{c}{\textbf{0.65}} \\
          & \textit{Collector} & 4     & 10    & 1     & 11.8  & 9.8   & 0     & 0     & 3.2   & 5     & 2     & 0     & 1.4   & 0     & 5     & 11.6  & \multicolumn{1}{c}{0.24} \\
          & Total & 15    & 18    & 6     & 26    & 28    & 6     & 12    & 12    & 31    & 6     & 15    & 8     & 0     & 10    & 50    &  -\\
    \midrule
    \multirow{4}[2]{*}{Coins collected} & \textit{Runner} & 1     & 2     & 0     & 0     & 0     & 4     & 4     & 5     & 0     & 0     & 1     & 0     & 2     & 1     & 0     & \multicolumn{1}{c}{0.07} \\
          & \textit{Killer} & 2.8   & 4     & 7     & 5.4   & 0     & 3     & 1.2   & 2.4   & 0     & 2.8   & 2     & 1     & 0     & 3     & 5.2   & \multicolumn{1}{c}{0.21} \\
          & \textit{Collector} & \textbf{13} & \textbf{12} & \textbf{4} & \textbf{13} & \textbf{4} & \textbf{8} & \textbf{20.6} & \textbf{22.8} & 0     & \textbf{8} & \textbf{14.4} & \textbf{3} & \textbf{7} & \textbf{3} & \textbf{14.2} & \multicolumn{1}{c}{\textbf{0.54}} \\
          & Total & 21    & 23    & 23    & 21    & 8     & 22    & 35    & 27    & 0     & 23    & 16    & 3     & 24    & 5     & 17    & -\\
    \midrule
    \multirow{3}[2]{*}{Time} & \textit{Runner} & \textbf{9} & \textbf{8} & -1    & 12    & \textbf{10} & -1    & \textbf{12} & \textbf{10}    & \textbf{10} & -1    & \text{13}    & \text{-1}     & \textbf{8} & \textbf{8}   & -1 & -\\
          & \textit{Killer} & 12    & 11    & -1    & -1    & 15    & -1    & 13    & -1    & 14    & -1    & \textbf{10} & 25    & -1    & 10    & -1    & -\\
          & \textit{Collector} & 12    & 11    & -1    & -1    & 13    & -1    & 16    & 15    & \textbf{10} & -1    & 13    & \textbf{22}    & -1    & 9     & -1    & -\\
    \bottomrule
    \end{tabular}%
  \label{Validation of agents}%
\end{table*}%

\subsubsection{Ability}

We denote the basic three persona agents without any modification as \textit{perfect} agents. A \textit{blind} agent is defined as the $A^*$ agent that can only search positions half distance to the current position than a \textit{perfect} agent. It means that the two agents own the same persona, but different abilities. We aim to generate levels with different skill-depth that require further planning and let the \textit{blind} agent fail to complete. The $Ability$ metric is formulated as:
\begin{equation}
Ability=
\left\{
\begin{aligned}
&-1  & p_{perfect}=1 \text{ and } p_{blind}<1, \\
&p_{blind} - p_{perfect} & otherwise,
\end{aligned}
\right.    
\end{equation}
\noindent where $p_{perfect}$ and $p_{blind}$ are the fractions of the level that the \textit{perfect} agent and \textit{blind} agent complete, respectively.

\subsubsection{Variance}
Diversity of levels are regarded as an important factor to improve the engagement. We focus on the diversity of events triggered by an agent during playing. The event type is same in the Table \ref{event}. We consider the coefficient of variance (CV) of the position of events, defined as the ratio of its standard deviation to its expectation: 
\begin{equation}\label{eq:variance}
CV(x) = \frac{Std(x)}{E(x)},    
\end{equation}
if considering an event's $x$-coordinate value.
We want to maximise the CV for both $x$- and $y$-coordinates of events in levels:
\begin{equation}
F=
\left\{
\begin{aligned}
&-p  & p < 1, \\
&-p - CV(x) - CV(y) & p = 1.
\end{aligned}
\right.    
\end{equation}
\noindent

\section{Experimental Study and Discussion}\label{sec:xp}

We first perform a preliminary experimental study to verify the behaviour preference of three agents on original SMB levels. 
Then, SMB levels are generated by searching in the latent spaces of a GAN generator with different evaluation metrics (cf. Section \ref{sec:designmetrics}) as the fitness function and different evaluation agents (cf. Section \ref{sec:agents}) for simulation. After level generation, all the levels are evaluated with various metrics based on their contents. Results of simulation-based tests by different playing agents are also reported.

\subsection{Validation of Agents}\label{sec:agentvalidation}
To verify if the designed agents (\textit{Runner}, \textit{Killer} and \textit{Collector}) act differently when they play the same levels, they are tested on the 15 original SMB levels. The time limit for the agent to determine an action is 200ms. We record the fraction of the level that was completed, the number of kills and times of collecting in each level. Every agent played each level 5 times because an agent's behaviour may vary when playing the same level twice. 

\begin{figure}[htbp]
\centering
\includegraphics[width=0.2\textwidth]{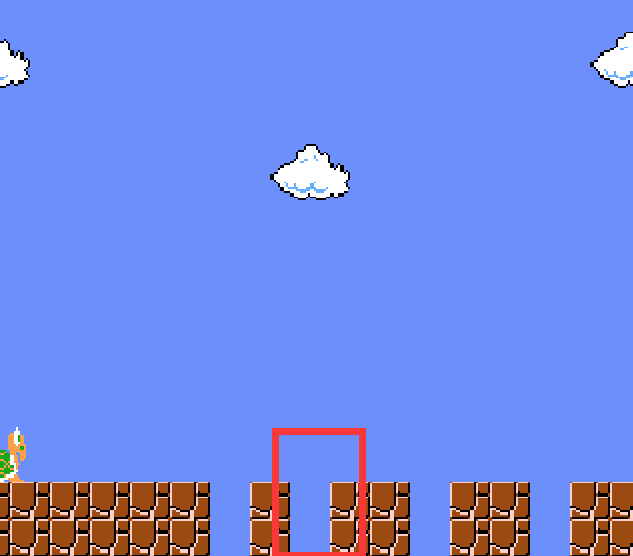}
\caption{\textit{Runner} falls into the gap, while the others don't.}
\label{agent fail case}
\end{figure}

Table \ref{Validation of agents} shows the ratio of completion, monsters killed, coins collected and completion time of the three agents. Comparing the ratio of completion, three agents show the similar ability to complete the levels. \textit{Runner} may pursue speed too much and choose a dangerous way in some levels, which leads to a low ratio of completion. As shown in Fig. \ref{agent fail case}, \textit{Runner} lands on the edge of the block and fails to jump again, while \textit{Killer} and \textit{Collector} choose a safer policy and successfully jump over these gaps. Table \ref{Validation of agents} also shows the significantly different preference of agents. In all the 15 levels, \textit{Killer} kills the most monsters, \textit{Collector} collects the most coins and \textit{Runner} finishes the levels with the minimum time. It proves that these agents have different behaviour preference when playing same levels.

\subsection{Generating Levels of Different Behaviour Engagement}\label{sec:levelgeneration}

With five evaluation metrics and three agents of different persona, we generate levels for each evaluation metric and for each agent. Hence, there are 15 groups of levels in total. Each group contains 30 level segments after evolving the latent vector with CMA-ES. The parameter setting of CMA-ES is same as in \cite{Volz2018Evolving}. The evaluation budget is 1,000, but 2,000 for metric \textit{Ability} because every evaluation requires two game simulations (i.e., one game each by the \emph{perfect} and \emph{blind} agents). The game engine used is the Mario AI framework\footnote{\url{https://github.com/amidos2006/Mario-AI-Framework}}. 



To answer the two research questions, two sets of level tests are performed to compare the generated levels based on their content and gameplay data of different persona agents.

\subsubsection{How different the generated levels are if using the same evaluation metric but different evaluation agent}

In order to test whether the levels generated with different agents have different features, we evaluate the generated levels on their contents. Table \ref{content test} shows the analysis on the generated levels. Each level is evaluated with the number of monsters, number of coins, number of gaps and the maximum width of gaps as those contents are the core elements in a SMB level. 

\begin{figure}[htbp]
\centering
\begin{subfigure}[b]{1\columnwidth}\centering
\includegraphics[width=.8\columnwidth]{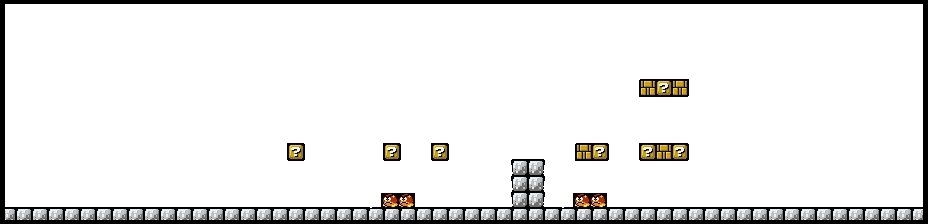}
\caption{\textit{Jump}.}
\end{subfigure}

\begin{subfigure}[b]{1\columnwidth}\centering
\includegraphics[width=.8\columnwidth]{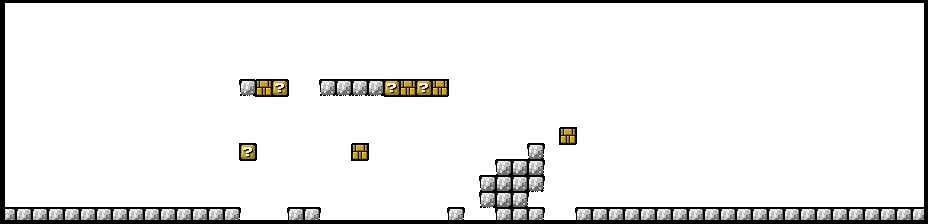}
\caption{\textit{Fail rate}.}
\end{subfigure}

\begin{subfigure}[b]{1\columnwidth}\centering
\includegraphics[width=.8\columnwidth]{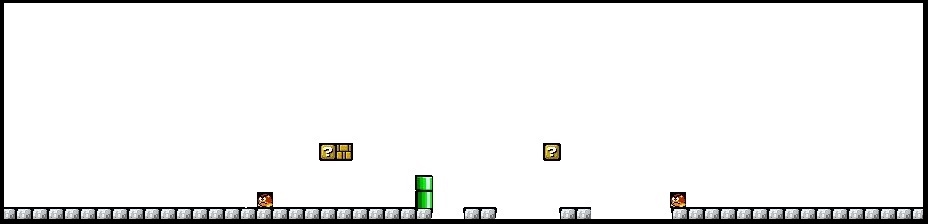}
\caption{\textit{Ability}.}
\end{subfigure}

\begin{subfigure}[b]{1\columnwidth}\centering
\includegraphics[width=.8\columnwidth]{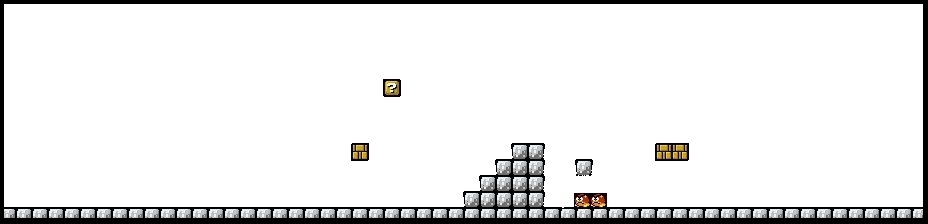}
\caption{\textit{Variance}.}
\end{subfigure}

\begin{subfigure}[b]{1\columnwidth}\centering
\includegraphics[width=.8\columnwidth]{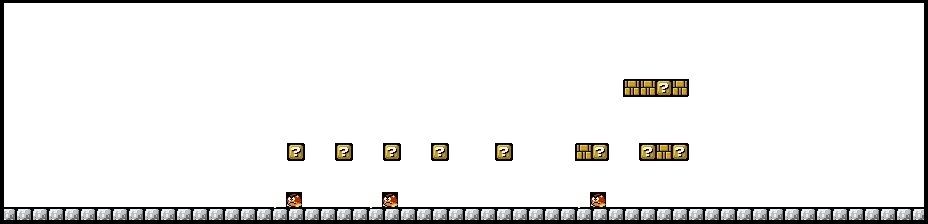}
\caption{\textit{Event}.}
\end{subfigure}
\caption{Levels generated with different evaluation metrics and evaluation agent \textit{Collector}. Levels generated with \textit{Jump} and \textit{Event} have a great amount of coins. \textit{Fail rate} and \textit{Ability} contain coins that near to the gap and dangerous to collect. In the level generated with \textit{Variance}, the position of coin is high and also increase the difficulty of collecting it.}
\label{content collector}
\end{figure}

Comparing the levels generated using different evaluation metrics and a same evaluation agent, as illustrated in Table \ref{content test}, it implies that game contents can be evolved in two ways: increasing element numbers and increasing difficulty. For instance, considering \textit{Collector} as the evaluation agent, \textit{Jump} and \textit{Event} metrics show the number of coins in the levels are significantly higher than the ones obtained using other evaluation agents. However, in other metrics, the number of coins shows no increase and sometimes lower than the ones obtained with other evaluation agents. But the number of gaps and max width of gaps in \textit{Fail rate} and \textit{Ability} are greater than that in \textit{Jump} and \textit{Event}. It indicates that \textit{Jump} and \textit{Event} metrics tend to evolve content in the first way, while \textit{Fail rate} and \textit{Ability} metrics tend to evolve content in the second way. Fig. \ref{content collector} shows this difference. Both two way of evolution can improve the engagement of \textit{Collector} and they are captured by different evaluation metrics.
 
Comparing \textit{Jump} and \textit{Event} metrics, it is interesting that the number of monsters increases significantly for \textit{Killer}. It indicates that the meaning of action (e.g., jump) is sometimes vague and may not guide the evolution as expected. For example, the purpose of jump can be killing monsters, dodging monsters, crossing gaps or even nothing. So this result shows that combining the action with game events can understand the player's behaviour purpose better.

\subsubsection{Considering a level generated with an evaluation agent, will a player that has similar persona gain more engagement compared with players that have another personas}
To study whether the generated levels can improve the behaviour engagement for particular personas, we test all the 15 groups of levels with three agents. Tables \ref{agent test: Jump} to \ref{agent test: Variance} show the results of agents playing the 15 groups of levels. Reading guidelines are given in the table captions.

\begin{figure}[htbp]
\centering
\includegraphics[width=.8\columnwidth]{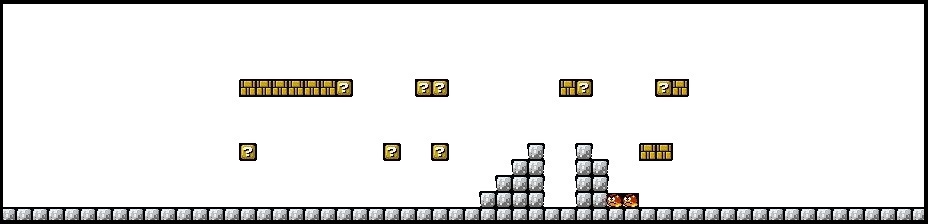}
\caption{A level generated by \textit{Event} metric and \textit{Collector} agent. It attracts \textit{Collector} to jump more for collection.}
\label{level_generated_collector}
\end{figure}

Considering \textit{Event} and \textit{Jump}, if fix the test agents, the test value like jump and number of events will be significantly higher when the test agent and the evaluation agent are the same. If fixing the evaluation agent and compare different test agents, more events are also triggered when the test agent and evaluation agent are the same. It indicates the generated levels have particular behaviour engagement indeed. For instance, Fig. \ref{level_generated_collector} illustrates a level generated using the \textit{Event} metric and \textit{Collector} as its evaluation agent. This level tends to attract \textit{Collector} to jump and collect coins, triggering more events. But for \textit{Runner} and \textit{Killer}, the presence of coins doesn't increase the appeal.

\textit{Fail rate} and \textit{Ability} metrics also can generate levels for specific persona. For \textit{Collector} as the evaluation agent, when test agent is \textit{Collector}, the completion time, jump, event type, number of events are all higher than the ones obtained by other test agents. When \textit{Killer} is used as the evaluation agent, the \textit{Fail rate} is higher when the test agent is also \textit{Killer}. This indicates that \textit{Fail rate} and \textit{Ability} can evolve levels that can be harder for specific persona, e.g., a coin hard to get or a monster hard to kill. It is consistent with the levels shown in Fig. \ref{content collector}.

\begin{figure}[htbp]
\centering
\includegraphics[width=.8\columnwidth]{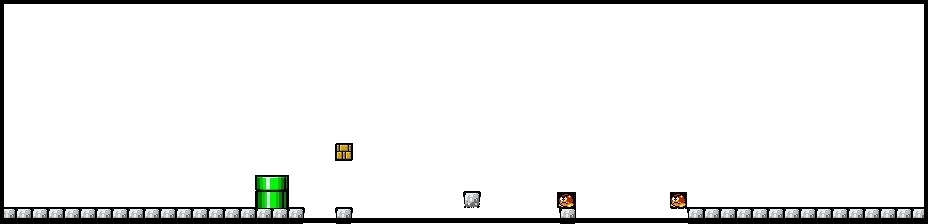}
\caption{A level generated by \textit{Fail rate} metric and \textit{Runner} agent. The width of gaps is large and consequently increase difficulty for all the three agents to pass the level.}
\label{level_generated_runner}
\end{figure}

However, when fixing the test agent and considering different evaluation agents, both \textit{Collector} and \textit{Killer} have much larger \textit{Fail rate} when \textit{Runner} is the evaluation agent, compared with the corresponding evaluation agent. This indicates that levels generated with \textit{Runner} as its evaluation agent can be also very challenging for \textit{Collector} and \textit{Killer}. From Table \ref{content test}, the Max width of gaps in \textit{FailRate-runner} and \textit{Ability-runner} is significantly higher. Fig. \ref{level_generated_runner} shows a level generated using the \textit{Fail rate} and \textit{Runner}. The agent needs to rely on several landing points to skip large interval segments. This level is very difficult for all the agents. \textit{Runner} and \textit{Collector} have 4.61\% and 5.47\% failure rates in this level. \textit{Killer} cannot even complete the level with \textit{Fail rate} 27.01\%. 

\begin{figure*}[htbp]
\centering
\begin{subfigure}[b]{.32\textwidth}\centering
\includegraphics[width=.8\textwidth]{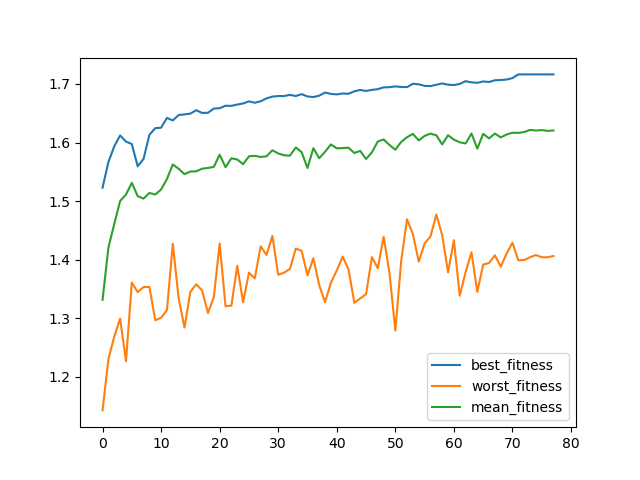}
\caption{\textit{Runner}.}
\end{subfigure}\hfill
\begin{subfigure}[b]{.32\textwidth}\centering
\includegraphics[width=.8\textwidth]{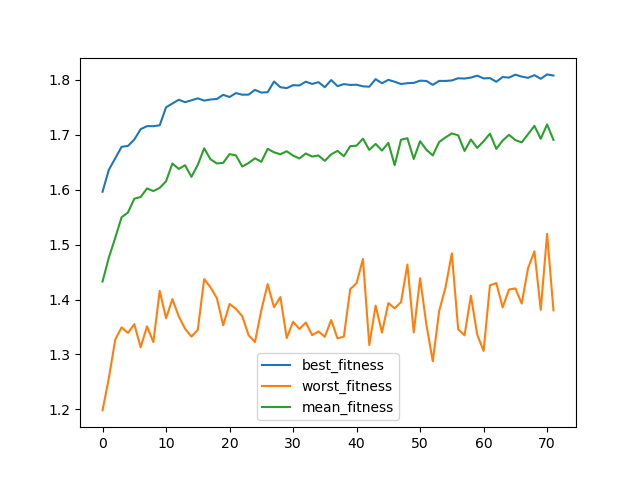}
\caption{\textit{Killer}.}
\end{subfigure}\hfill
\begin{subfigure}[b]{.32\textwidth}\centering
\includegraphics[width=.8\textwidth]{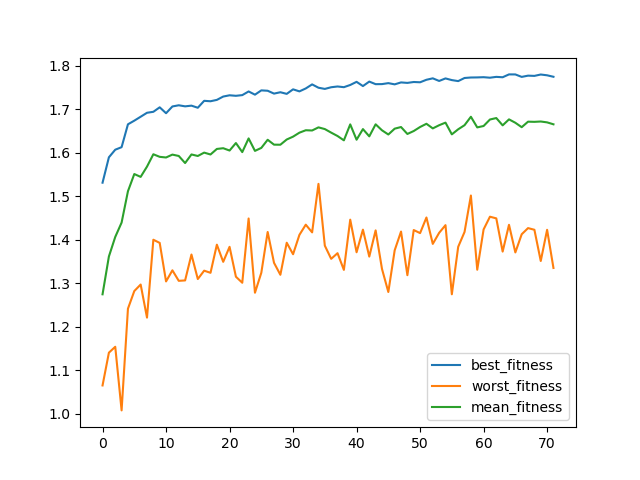}
\caption{\textit{Collector}.}
\end{subfigure}
\caption{Convergence curves of the best (blue), average (green) and worst (yellow) \textit{variance} (Eq. \ref{eq:variance}) values over 30 trials.}
\label{fitness}
\end{figure*}

For \textit{Variance}, the pattern is not clear and no obvious preference observed from the results. A possible reason is that \textit{Variance} considers all the types of content together, while in many cases multiple events happen closely, such like jump and collect. Some meaningless event should also be excluded during evaluation. 
There is an extreme fail case for \textit{Variance}. There is nothing but ground in the level, while \textit{Collector} will jump time to time and increase the \textit{Variance} on x-axis. Fig. \ref{fitness} shows the premature convergence of fitness. 

\def\removed{

\begin{figure*}[htbp]
\begin{subfigure}[b]{.32\textwidth}
\includegraphics[width=1\textwidth]{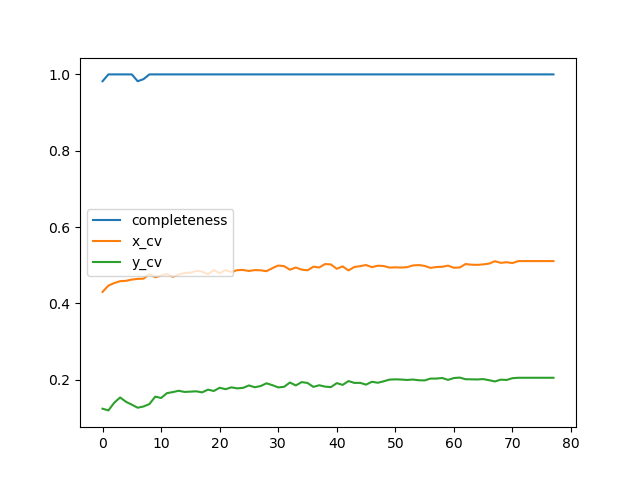}
\caption{\textit{Runner}.}
\end{subfigure}\hfill
\begin{subfigure}[b]{.32\textwidth}
\includegraphics[width=1\textwidth]{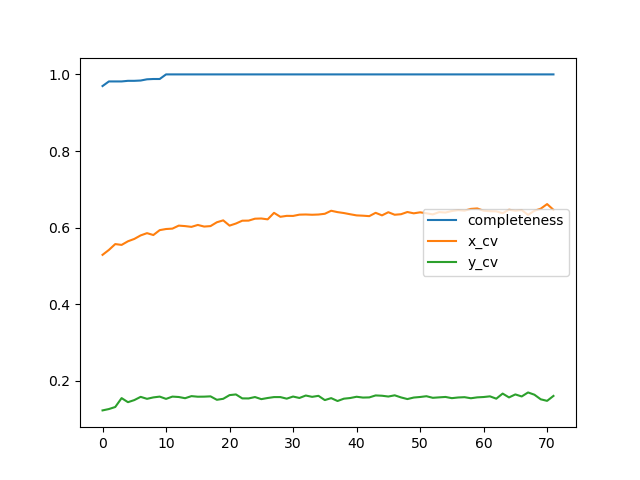}
\caption{\textit{Killer}.}
\end{subfigure}\hfill
\begin{subfigure}[b]{.32\textwidth}
\includegraphics[width=1\textwidth]{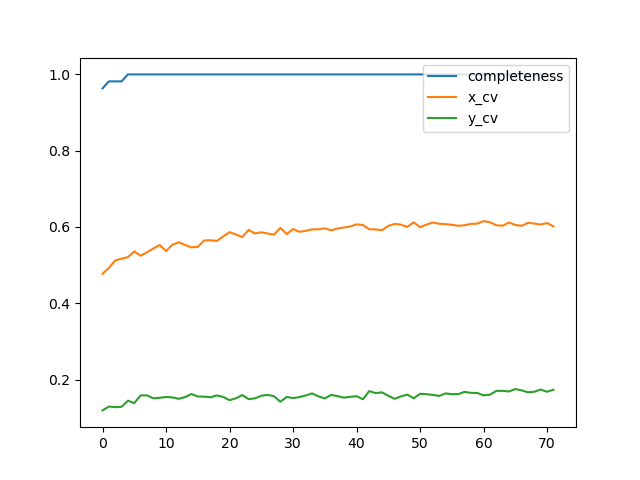}
\caption{\textit{Collector}.}
\end{subfigure}
\caption{Values of terms in the \textit{Variance} metric. Blue: ratio of completeness; green/yellow: \textit{Variance} of y-axis/x-axis of the best solution of each iteration. All the values are averaged over 30 independent trials.}
\label{Separate}
\end{figure*}
}
 
 \def\removed{
\begin{figure}[htbp]
\centering
\includegraphics[width=.8\columnwidth]{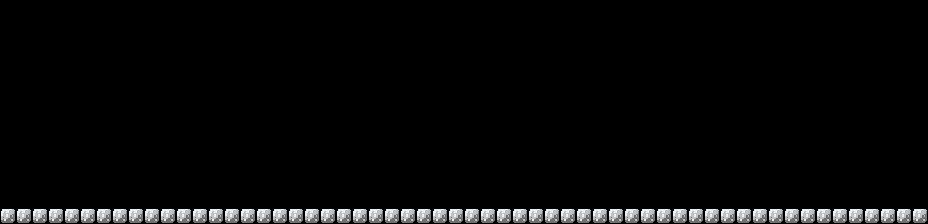}
\caption{A boring level generated by metric \textit{Variance} and \textit{Collector} as evaluation agent. There is nothing except ground.}
\label{fail case}
\end{figure}}


\begin{table*}[htbp]
  \centering
  \caption{Content analysis on levels generated using different evaluation agents and metrics. The first column gives the evaluation agent and evaluation metric. ``max width'' is the maximum width of gap in a level. Bold number is the maximum value compared with levels generated with three agents but the same evaluation metric. ``+''/``-'' indicate that the value of \textit{Killer} or \textit{Collector} is significantly larger/lower than that of \textit{Runner} according to Wilcoxon rank-sum test with $p<0.05$.}
    \begin{tabular}{l|cc|cc|cc|cc}
    \toprule
    \textbf{Metric-Agent} & \textbf{Avg \#monsters} & \textbf{Std \#monsters} & \textbf{Avg \#coins} & \textbf{Std \#coins} & \textbf{Avg \#gap} & \textbf{Std \#gap} & \textbf{Avg max width} & \textbf{Std max width} \\
    \midrule
    \textit{Jump-Runner} & 1.97  & 1.11  & 0.5 & 1.26  & 0.7   & 1.49  & \textbf{1.47}  & 4.03 \\
    \textit{Jump-Killer} & 1.77  & 1.26  & 1.2 & 1.89  & \textbf{0.83}  & 1.61  & 1.13  & 2.28 \\
    \textit{Jump-Collector} & \textbf{2.43}  & 1.54  & \textbf{4.63+} & 2.9   & 0.73  & 1.12  & 1.23  & 2.01 \\
    \midrule
    \textit{Fail rate-Runner} & 3     & 1.86  & 1.67 & 1.66  & 1.13  & 1.26  & \textbf{6.07}  & 7.42 \\
    \textit{Fail rate-Killer}& 3.03  & 2.81  & 1.67 & 1.83  & \textbf{2.27+} & 1.88  & 4.37  & 3.84 \\
    \textit{Fail rate-Collector} & \textbf{3.2}   & 3.25  & \textbf{2.5} & 2     & 1.97  & 1.94  & 3.67  & 4.26 \\
    \midrule
    \textit{Ability-Runner} & 0.93  & 1.06  & 0.7 & 1.22  & \textbf{2.83}  & 1.04  & \textbf{8.8}  & 4.55 \\
    \textit{Ability-Killer} & 0.87  & 1.06  & \textbf{1.53+} & 1.67  & 2.77  & 1.17  & 5.67- & 3.03 \\
    \textit{Ability-Collector} & \textbf{1.1}   & 1.22  & 1.2 & 1.62  & 2.33  & 1.25  & 5.3- & 4.27 \\
    \midrule
    \textit{Variance-Runner} & \textbf{2.17}  & 1.16  & \textbf{2.13} & 2.46  & 0.53  & 0.99  & \textbf{1.83}  & 4.45 \\
    \textit{Variance-Killer} & 1.87  & 0.72  & 1.77 & 1.28  & \textbf{0.83} & 1.24  & 1.17  & 1.93 \\
    \textit{Variance-Collector} & 1.5- & 0.92  & 1.03 & 0.84  & 0.03  & 0.18  & 0.07  & 0.36 \\
    \midrule
    \textit{Event-Runner} & 2.87  & 1.5   & 2.37 & 2.6   & \textbf{0.53}  & 0.96  & \textbf{1.6}   & 3.94 \\
    \textit{Event-Killer} & \textbf{4.23+} & 1.71  & 2.93 & 2.74  & 0.5   & 1.15  & 0.67  & 1.62 \\
    \textit{Event-Collector} & 2.53  & 1.73  & \textbf{5.67+} & 2.91  & 0.3   & 0.94  & 0.37  & 0.95 \\
    \bottomrule
    \end{tabular}%
  \label{content test}%
\end{table*}%

\begin{table*}[htbp]
  \centering
  \caption{\label{tab:behaviourtest}Agent behaviour test. \textit{Kill rate} and \textit{Collect rate} are defined in Section \ref{sec:designmetrics}.
  \textit{Time} means the completion time. 
  \textit{Event type} and \textit{Event num} means the type and number of events triggered by agent, respectively. \textit{Complete} means the ratio of completeness. All the values are averaged over 30 independent levels. The bold number represents the largest number (smallest for complete) for the same test agent and different evaluation agent. ``+''/``-'' represents the value is significantly larger/smaller than the baseline according to Wilcoxon rank-sum test with $p<0.05$. The baseline has the identical test and evaluation agent.}
 \begin{subtable}[h]{1\textwidth}
 \centering
 \caption{\label{agent test: Jump}Levels generated using \textit{Jump} metric and different evaluation agents.} 
    \begin{tabular}{p{8.065em}p{6.375em}cccccccc}
    \toprule
    \multicolumn{2}{c}{\textbf{Gameplay agent}}
     & \multicolumn{1}{p{4.19em}}{\textbf{Kill rate}} & \multicolumn{1}{p{5em}}{\textbf{Collect rate}} & \multicolumn{1}{p{5em}}{\textbf{Fail rate}} & \multicolumn{1}{p{4em}}{\textbf{Time}} & \multicolumn{1}{p{4em}}{\textbf{\#Jump}} & \multicolumn{1}{p{5em}}{\textbf{Event type}} & \multicolumn{1}{p{5em}}{\textbf{Event num}} & \multicolumn{1}{p{4em}}{\textbf{Complete}} \\
    \textbf{During generation}  & \textbf{During test} &&&&&&&&\\
    \midrule
    \textit{Collector} & \textit{Collector} & 0.27  & \textbf{0.78} & 1.29  & 3.83  & \textbf{8.47} & \textbf{4.43} & \textbf{21.47} & 1 \\
    \textit{Runner} & \textit{Collector} & \textbf{0.32}  & 0.07- & \textbf{2.11} & \textbf{4.8} & 6.1- & 3.73- & 13.1- & \textbf{0.95} \\
    \textit{Killer} & \textit{Collector} & 0.3   & 0.25- & 1.03  & 3.2- & 5.97- & 3.87- & 13.27- & 1 \\
    \midrule
    \textit{Collector} & \textit{Runner} & \textbf{0.07}  & \textbf{0.09+} & 0.12  & 2.3- & 2.77- & \textbf{3.57+} & 7.2- & \textbf{1} \\
    \textit{Runner} & \textit{Runner} & 0.01  & 0.01  & \textbf{0.58} & \textbf{3.4} & \textbf{6.37} & 3.07  & \textbf{13.8} & \textbf{1} \\
    \textit{Killer} & \textit{Runner} & \textbf{0.07}  & 0.03  & 0.14  & 2.5- & 3.67- & 3.17  & 8.53- & \textbf{1} \\
    \midrule
    \textit{Collector} & \textit{Killer} & \textbf{0.88} & \textbf{0.11}  & 0.93  & 3.07  & 4.23- & \textbf{4.43+} & 11.5- & 1 \\
    \textit{Runner} & \textit{Killer} & 0.77  & 0     & \textbf{4.82} & 3.03  & 5.03  & 3.67  & 11.87 & \textbf{0.93} \\
    \textit{Killer} & \textit{Killer} & 0.7   & 0.07  & 1.03  & \textbf{4.2} & \textbf{8.3} & 3.87  & \textbf{18.37} & 0.99 \\
    \bottomrule
    \end{tabular}%
    \end{subtable}
        
    \begin{subtable}[h]{1\textwidth}
    \centering
            \caption{\label{agent test: Event}Levels generated using \textit{Event} metric and different evaluation agents.}
    \begin{tabular}{p{8.065em}p{6.375em}cccccccc}
    \toprule
    \multicolumn{2}{c}{\textbf{Gameplay agent}}
     & \multicolumn{1}{p{4.19em}}{\textbf{Kill rate}} & \multicolumn{1}{p{5em}}{\textbf{Collect rate}} & \multicolumn{1}{p{5em}}{\textbf{Fail rate}} & \multicolumn{1}{p{4em}}{\textbf{Time}} & \multicolumn{1}{p{4em}}{\textbf{\#Jump}} & \multicolumn{1}{p{5em}}{\textbf{Event type}} & \multicolumn{1}{p{5em}}{\textbf{Event num}} & \multicolumn{1}{p{4em}}{\textbf{Complete}} \\
    \textbf{During generation}  & \textbf{During test} &&&&&&&&\\
    \midrule
    \textit{Collector} & \textit{Collector} & 0.46  & \textbf{0.86} & 0.64  & \textbf{5.6} & \textbf{8.47} & \textbf{4.6} & \textbf{23.17} & \textbf{0.98} \\
    \textit{Runner} & \textit{Collector} & \textbf{0.47}  & 0.57  & \textbf{2.18} & 3.07- & 5.67- & 4.47  & 14.33- & \textbf{0.98} \\
    \textit{Killer} & \textit{Collector} & 0.31  & 0.55- & 0.87  & 5- & 6.6- & 4.37  & 16.2- & \textbf{0.98} \\
    \hline
    \textit{Collector} & \textit{Runner} & 0.07  & 0.17  & 0.11  & 2.2- & 2.77- & 3.77  & 7.83- & \textbf{1} \\
    \textit{Runner} & \textit{Runner} & \textbf{0.16}  & \textbf{0.34}  & \textbf{0.54} & \textbf{2.9} & \textbf{5.23} & \textbf{4.07} & \textbf{13} & \textbf{1} \\
    \textit{Killer} & \textit{Runner} & 0.13  & 0.09- & 0.15  & 2.43- & 3.2- & 3.73  & 8.5- & \textbf{1} \\
    \hline
    \textit{Collector} & \textit{Killer} & 0.84  & 0.2   & 1.42  & 2.97- & 3.9- & 4.53  & 11.33- & 0.98 \\
    \textit{Runner} & \textit{Killer} & 0.8   & 0.24  & \textbf{2.55} & 2.93- & 4.8   & 4.4- & 12.87- & \textbf{0.97} \\
    \textit{Killer} & \textit{Killer} & \textbf{0.92} & \textbf{0.25}  & 0.72  & \textbf{3.93} & \textbf{6.27} & \textbf{4.63} & \textbf{17.63} & 1 \\
    \bottomrule
    \end{tabular}
    \end{subtable}
    
    \begin{subtable}[h]{1\textwidth}
        \centering
        \caption{\label{agent test: Fail Rate}Levels generated using \textit{Fail rate} metric and different evaluation agents. }
    \begin{tabular}{p{8.065em}p{6.375em}cccccccc}
    \toprule
    \multicolumn{2}{c}{\textbf{Gameplay agent}}
     & \multicolumn{1}{p{4.19em}}{\textbf{Kill rate}} & \multicolumn{1}{p{5em}}{\textbf{Collect rate}} & \multicolumn{1}{p{5em}}{\textbf{Fail rate}} & \multicolumn{1}{p{4em}}{\textbf{Time}} & \multicolumn{1}{p{4em}}{\textbf{\#Jump}} & \multicolumn{1}{p{5em}}{\textbf{Event type}} & \multicolumn{1}{p{5em}}{\textbf{Event num}} & \multicolumn{1}{p{4em}}{\textbf{Complete}} \\
    \textbf{During generation}  & \textbf{During test} &&&&&&&&\\
    \midrule
    \textit{Collector} & \textit{Collector} & 0.16  & 0.41  & 6.87  & \textbf{4.9} & \textbf{8.23} & \textbf{4.1} & \textbf{18.73} & 0.93 \\
    \textit{Runner} & \textit{Collector} & 0.26  & \textbf{0.54} & \textbf{14.33+} & 3.33- &4.5- & \textbf{4.1} & 11.2- & \textbf{0.84} \\
    \textit{Killer} & \textit{Collector} & 0.19  & 0.38  & 7.95  & 3.67- & 5.17- & 3.93  & 12.03- & 0.93 \\
    \hline
    \textit{Collector} & \textit{Runner} & 0.08  & 0.06  & 2.14  & 2.27  & 2.87  & 3.43  & 7.37  & 0.98 \\
    \textit{Runner} & \textit{Runner} & 0.16  & 0.08  & \textbf{7.37} & \textbf{2.33} & \textbf{3.13} & \textbf{3.53} & \textbf{7.97} & \textbf{0.93} \\
    \textit{Killer} & \textit{Runner} & 0.06  & 0.01  & 0.46- & 2.27  & 2.9   & 3.27  & 7.2   & 1 \\
    \hline
    \textit{Collector} & \textit{Killer} & 0.44  & 0.08  & 8.36  & \textbf{2.93} & \textbf{3.93} & 3.77  & \textbf{10.57} & 0.92 \\
    \textit{Runner} & \textit{Killer} & \textbf{0.74} & 0.23+ & \textbf{13.19+} & 2.57  & 3.33  & \textbf{4.23+} & 9.87  & \textbf{0.84} \\
    \textit{Killer} & \textit{Killer} & 0.48  & 0.06  & 9.26  & 2.8   & 3.77  & 3.67  & 9.8   & 0.91 \\
    \bottomrule
    \end{tabular}
  \end{subtable}
      
    \begin{subtable}[h]{1\textwidth}
  \centering
        \caption{  \label{agent test: ability}Levels generated using \textit{Ability} metric and different evaluation agents. }
    \begin{tabular}{p{8.065em}p{6.375em}cccccccc}
    \toprule
    \multicolumn{2}{c}{\textbf{Gameplay agent}}
     & \multicolumn{1}{p{4.19em}}{\textbf{Kill rate}} & \multicolumn{1}{p{5em}}{\textbf{Collect rate}} & \multicolumn{1}{p{5em}}{\textbf{Fail rate}} & \multicolumn{1}{p{4em}}{\textbf{Time}} & \multicolumn{1}{p{4em}}{\textbf{\#Jump}} & \multicolumn{1}{p{5em}}{\textbf{Event type}} & \multicolumn{1}{p{5em}}{\textbf{Event num}} & \multicolumn{1}{p{4em}}{\textbf{Complete}} \\
    \textbf{During generation}  & \textbf{During test} &&&&&&&&\\
    \midrule
    \textit{Collector} & \textit{Collector} & 0.1   & \textbf{0.38} & 2.97  & \textbf{2.53} & \textbf{4.6} & \textbf{3.6} & \textbf{10.07} & 0.98 \\
    \textit{Runner} & \textit{Collector} & 0.06  & 0.18  & \textbf{12.7+} & 2.37  & 3.97  & 3.33  & 8.33  & \textbf{0.89} \\
    \textit{Killer} & \textit{Collector} & \textbf{0.12}  & 0.3   & 4.5   & 2.47  & 4.2   & \textbf{3.6} & 9.13  & 0.97 \\
     \midrule
    \textit{Collector} & \textit{Runner} & \textbf{0.07}  & \textbf{0.03}  & 0.23- & 2.13  & 2.6   & \textbf{3.2} & 6.4   & 1 \\
    \textit{Runner} & \textit{Runner} & 0.04  & 0     & 0.58  & \textbf{2.2} & \textbf{2.83} & 3.1   & \textbf{6.77} & 1 \\
    \textit{Killer} & \textit{Runner} & 0.04  & 0     & \textbf{1.58} & 2.07  & 2.3- & 3.07  & 5.63- & \textbf{0.98-} \\
    \midrule
    \textit{Collector} & \textit{Killer} & 0.32  & \textbf{0.04}  & 5.41  & \textbf{2.47} & \textbf{3.87} & \textbf{3.43} & \textbf{8.47} & 0.97 \\
    \textit{Runner} & \textit{Killer} & 0.28  & 0     & \textbf{15.57+} & 2.2   & 3.43  & 3.3   & 7.27  & \textbf{0.84-} \\
    \textit{Killer} & \textit{Killer} & \textbf{0.35} & 0.01  & 5.33  & 2.43  & 3.43  & 3.4   & 7.43  & 0.97 \\
    \bottomrule
    \end{tabular}%
    \end{subtable}
        
    \begin{subtable}[h]{1\textwidth}
    \centering
    \caption{\label{agent test: Variance}Levels generated using \textit{Variance} metric and different evaluation agents. }
   \begin{tabular}{p{8.065em}p{6.375em}cccccccc}
    \toprule
    \multicolumn{2}{c}{\textbf{Gameplay agent}}
     & \multicolumn{1}{p{4.19em}}{\textbf{Kill rate}} & \multicolumn{1}{p{5em}}{\textbf{Collect rate}} & \multicolumn{1}{p{5em}}{\textbf{Fail rate}} & \multicolumn{1}{p{4em}}{\textbf{Time}} & \multicolumn{1}{p{4em}}{\textbf{\#Jump}} & \multicolumn{1}{p{5em}}{\textbf{Event type}} & \multicolumn{1}{p{5em}}{\textbf{Event num}} & \multicolumn{1}{p{4em}}{\textbf{Complete}} \\
    \textbf{During generation}  & \textbf{During test} &&&&&&&&\\
    \midrule
    \textit{Collector} & \textit{Collector} & \textbf{0.45}  & 0.56  & 0.13  & 2.47  & 3.7   & \textbf{4.37} & 8.83  & 1 \\
    \textit{Runner} & \textit{Collector} & \textbf{0.45}  & 0.46  & \textbf{3.84+} & 2.9+ & \textbf{5.2+} & 4.13+ & \textbf{13.2} & 0.97 \\
    \textit{Killer} & \textit{Collector} & 0.15- & \textbf{0.66} & 1.01+ & \textbf{4.5} & 4.4+ & 3.9- & 10.43+ & \textbf{0.96} \\
    \midrule
    \textit{Collector} & \textit{Runner} & \textbf{0.14}  & \textbf{0.1}   & 0.05  & 2.1- & 1.83- & \textbf{3.27} & 5.07  & \textbf{1} \\
    \textit{Runner} & \textit{Runner} & 0.09  & 0     & \textbf{0.63} & \textbf{2.4} & \textbf{2.7} & 3.17  & \textbf{6.63} & \textbf{1} \\
    \textit{Killer} & \textit{Runner} & 0     & 0     & 0.06  & 2.13  & 2- & 3     & 5- & \textbf{1} \\
    \midrule
    \textit{Collector} & \textit{Killer} & 0.81  & \textbf{0.35+} & 0.09- & 2.8   & 3.13  & \textbf{4.23} & 8.17  & 1 \\
    \textit{Runner} & \textit{Killer} & 0.8   & 0.13  & \textbf{2.75} & \textbf{2.9+} & \textbf{3.77+} & 4.13  & \textbf{10.1+} & \textbf{0.97} \\
    \textit{Killer} & \textit{Killer} & \textbf{0.94} & 0.04  & 0.35  & 2.57  & 2.67  & 4.07  & 7.13  & 1 \\
    \bottomrule
    \end{tabular}%
  \end{subtable}
\end{table*}%


\section{Conclusion \& Future work}\label{sec:conclusion}

In this paper, we first analyse the metrics for evaluating game levels. Then, this work uses evaluation metrics that have no preference bias and designed agents with different personas to investigate whether the levels generated with different simulation agent can have different behaviour engagement. Experimental results on a platformer game show that our framework can adapt to different personas and generate levels for them specifically via changing the behaviour preference of evaluation agents. The work may guide game designers and researchers to study further on combining these different evaluation metrics to generate levels with more possibilities (e.g., dynamically changing the engagement for general players or particularly for players of a certain persona).

In this work, we assume that higher number (difficulty, diversity) of triggered events represents more engagement of players. A more reasonable way is to keep them within a range, neither too boring nor too diverse. One important future work is conducting human test to further verify whether those generated levels have different behaviours engagement. The evaluation agents can also be a combination of personas (e.g., combining Collector and Killer) or a machine learning based agent to imitate players better. It is also interesting to test our system on other game genres.

\section*{Acknowledgement}
The authors would like to thank the anonymous reviewers for their valuable comments.

\bibliographystyle{IEEEtran}
\balance
\bibliography{engagementmetrics}


\end{document}